\documentclass[11pt,twoside,twocolumn,a4paper]{article}

\usepackage{cvww}
\usepackage{times}
\usepackage{epsfig}
\usepackage{graphicx}
\usepackage{amsmath}
\usepackage{amssymb}
\usepackage{caption}
\usepackage{subcaption}
\usepackage{multirow}

\usepackage{rotating}
\usepackage{float}

\usepackage{cuted}
\usepackage{capt-of}

\usepackage{cleveref}

\cvwwfinalcopy 

\usepackage[printwatermark]{xwatermark}
\newwatermark*[allpages,color=magenta,angle=0,scale=1,xpos=0,ypos=400pt,fontsize=6pt]{
  The content of this paper was published in CVWW, 2020. This ArXiv version was published after the peer review. Please, cite the following paper:\\
  Jaka {\v{S}}ircelj, Tim Oblak, Klemen Grm, Uro{\v{s}} Petkovi{\'{c}}, Ale{\v{s}} Jakli{\v{c}}, Peter Peer, Vitomir {\v{S}}truc and Franc Solina:\\
  "Segmentation and Recovery of Superquadric Models using Convolutional Neural Networks", 25th Computer Vision Winter Workshop, 2020
}

\def\cvwwPaperID{34} 

\ifcvwwfinal\fi
\begin{document}

\title{Segmentation and Recovery of Superquadric Models using Convolutional Neural Networks}

\author{\normalsize{Jaka {\v{S}}ircelj$^{1,2}$, Tim Oblak$^2$, Klemen Grm$^1$, Uro{\v{s}} Petkovi{\'{c}}$^1$,}\\
\normalsize{Ale{\v{s}} Jakli{\v{c}}$^2$, Peter Peer$^2$, Vitomir {\v{S}}truc$^1$ and Franc Solina$^2$}\\
\normalsize{$^1$ Faculty of Electrical Engineering, UL, Tr{\v{z}}a{\v{s}}ka 25, Ljubljana, Slovenia}\\
\normalsize{$^2$ Faculty of Computer and Information Science, UL, Ve{\v{c}}na pot 113, Ljubljana, Slovenia}\\
{\tt\small jaka.sircelj@fe.uni-lj.si}
\vspace{-1cm}
}
\maketitle

\ifcvwwfinal\thispagestyle{fancy}\fi

\begin{strip}
    \centering
    {\includegraphics[width=0.85\linewidth]{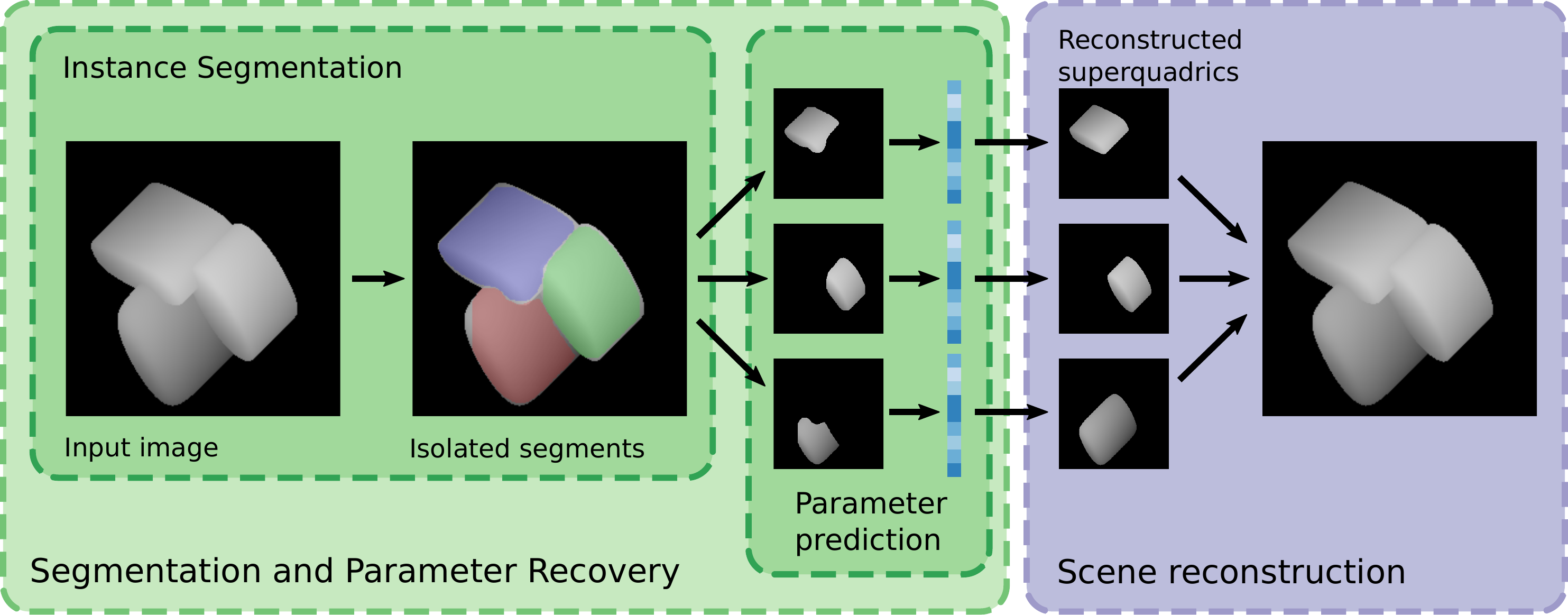}}
    \captionof{figure}{We study the problem of segmenting and recovering superquadric models from depth scenes. Our approach uses instance segmentation with Mask-RCNNs followed by superquadric-parameter estimation from incomplete data with a standard CNN (left part of the figure). Using the recovered superquadric models we are able to efficiently reconstruct the original depth scene (right part of the figure).}
    \label{fig:header}
\end{strip}



\begin{abstract}
   In this paper we address the problem of representing 3D visual data with parameterized volumetric shape primitives. Specifically, we present a (two-stage) approach built around convolutional neural networks (CNNs) capable of segmenting complex depth scenes into the simpler geometric structures that can be represented with superquadric models. In the first stage, our approach uses a Mask-RCNN model to identify superquadric-like structures in depth scenes and then fits superquadric models to the segmented structures using a specially designed CNN regressor. Using our approach we are able to describe complex structures with a small number of interpretable parameters. We evaluated the proposed approach on synthetic as well as real-world depth data and show that our solution does not only result in competitive performance in comparison to the state-of-the-art, but is able to decompose scenes into a number of superquadric models at a fraction of the time required by competing approaches. We make all data and models used in the paper available from \footnotesize{\url{https://lmi.fe.uni-lj.si/en/research/resources/sq-seg}}.\vspace{-6mm}
\end{abstract}


\section{Introduction}

Representing three-dimensional visual data in terms of parameterized shape primitives represents a longstanding goal in computer vision. The interest in this problem is fueled by the vast number of applications that rely on concise descriptions of the physical 3D space in various sectors ranging from autonomous driving and robotics to space exploration, medical imaging and beyond~\cite{driving_Levinson, mars_objects, 3Dxray}.

Past research in this area has looked at different models that could act as volumetric shape primitives, such as generalized cylinders~\cite{Zhou2015gencylinders} or cuboids~\cite{cuboids_tulsiani, niu2018cuboidsseg,jiang2013cuboids}, but superquadrics established themselves as one of the most suitable choices for this task~\cite{bajcsy1987three,solina1990recovery,jaklic2000,slabanja2018sq, oblak2019recovery, paschalidou2019superquadrics} due to their ability to represent a wide variety of 3D shapes, such as ellipsoids, cylinders, parallelopipeds and various shapes in between. Formally, superquadrics are defined by an implicit 3D closed surface equation, i.e.:\vspace{-5mm}

{\small
\begin{equation}
   \Bigg(\bigg(\frac{x - x_0}{a_1}\bigg)^{\frac{2}{\epsilon_2}}\hspace{-2.8mm}+ \bigg(\frac{y - y_0}{a_2}\bigg)^{\frac{2}{\epsilon_2}}\Bigg)^{\frac{\epsilon_2}{\epsilon_1}}\hspace{-2.8mm}+ \Bigg(\frac{z - z_0}{a_3}\Bigg)^{\frac{2} {\epsilon_1}} \hspace{-2mm}=1
    \label{eq:sq_impl}
\end{equation}
}
\noindent
\hspace{-0.18cm}where $a_1, a_2, a_3$ define the bounding box size of the superquadric, $\varepsilon_1$ and $\varepsilon_2$ define it's shape and $(x_0, y_0, z_0)^\intercal$ represent the center of the superquadric in a reference coordinate system \cite{jaklic2000}. Existing techniques for recovering superquadric models typically involve costly iterative parameter-estimation procedures that further increase in complexity if more than a single superquadric needs to be fitted to a scene~\cite{leonardis1997superquadrics, jaklic2000}. With complex scene geometries, superquadric recovery must necessarily be combined with segmentation techniques capable of partitioning the scene into simpler superquadric-like structures. This, however, puts a considerable computational burden on the fitting procedure as state-of-the-art techniques for recovery-and-segmentation of multiple superquadric models are typically extremely resource demanding.

With recent advances in computer vision and more importantly deep learning, it is possible to design solutions for simultaneous segmentation and recovery of superquadrics that are much more efficient than existing solutions. In this paper, we, therefore, revisit the problem of representing complex depth scenes with multiple superquadrics and develop an efficient solution for this task around convolutional neural networks (CNNs). Specifically, we assume that small superquadric-like structures in range images can be modeled as instances of a specific class of objects, and, therefore, train a Mask-RCNN~\cite{MaskRCNN} model to segment the scene, as illustrated in Fig.~\ref{fig:header}. The results of this instance segmentation are then used as input to a second CNN that recovers superquadric parameters for each of the identified superquadric-like objects. Because the identified superquadric-like objects may be partially occluded, we account for this fact during training and learn the parameters of the second CNN in a robust manner. We evaluate the performance of our approach on simulated, but also real-world range images. We achieve segmentation and recovery results comparable to the state-of-the-art, but achieve a considerable speed-up, which makes the developed solution suitable for a much wider range of applications. We note that in this paper we approach a constrained superquadric recovery problem, where we assume that the depth scene can be approximated by a number of unrotated superquadric models. 

Our main contributions in this paper are:\vspace{-0.5mm}
\begin{itemize}
    \item We present a novel solution for segmentation and recovery of multiple (unrotated) superquadric models from range images built around CNNs and evaluate it in experiments with simulated and real-world depth data.\vspace{-2mm}
    \item We show that existing Mask-RCNNs may be used for identifying superquadric-like structures in range images in an efficient manner.\vspace{-2mm}
    \item We demonstrate that superquadrics can be recovered from partial depth data using a simple CNN-based regressor and the parameter estimation errors are comparable to the error produced by state-of-the-art techniques used for this task.  
\end{itemize}



\section{Related work}\label{Sec: related}

Existing techniques to scene segmentation with superquadrics can in general be divided in one of two groups: \textit{i)} techniques that approach the problem by segmenting the scene and recovering superquadrics at the same time (\textit{segment-and-fit}), and \textit{ii)} techniques that first segment the scene and then fit superquadric models to the segmented parts (\textit{segment-then-fit}). In this section we briefly review both groups of techniques with the goal of providing the necessary context for our work. For a more comprehensive coverage of the subject, the reader is referred to~\cite{jaklic2000}.
\begin{figure*}[tb]
    \centering
    \newcommand{\datspace}{0.0953\linewidth}
    \begin{subfigure}[t]{\datspace}
        \includegraphics[width=\textwidth]{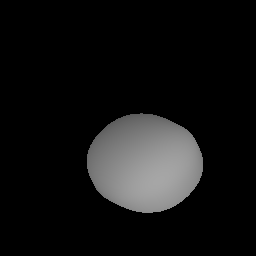}
    \end{subfigure}
    \hfill
    \begin{subfigure}[t]{\datspace}
        \includegraphics[width=\textwidth]{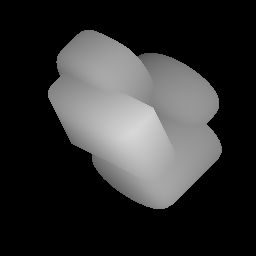}
    \end{subfigure}
    \hfill
    \begin{subfigure}[t]{\datspace}
        \includegraphics[width=\textwidth]{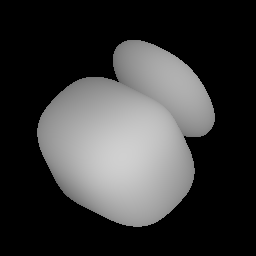}
    \end{subfigure}
    \hfill
    \begin{subfigure}[t]{\datspace}
        \includegraphics[width=\textwidth]{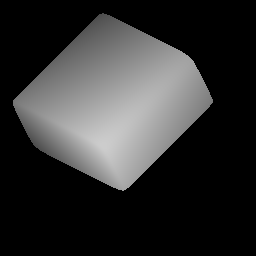}
    \end{subfigure}
    \hfill
    \begin{subfigure}[t]{\datspace}
        \includegraphics[width=\textwidth]{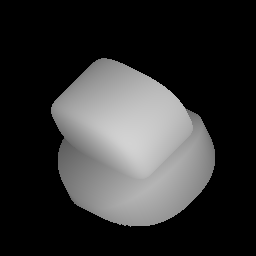}
    \end{subfigure}
    \hfill
    \begin{subfigure}[t]{\datspace}
        \includegraphics[width=\textwidth]{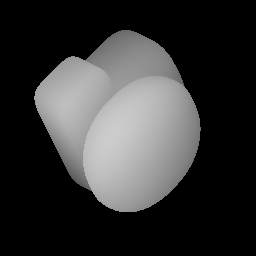}
    \end{subfigure}
    \hfill
    \begin{subfigure}[t]{\datspace}
        \includegraphics[width=\textwidth]{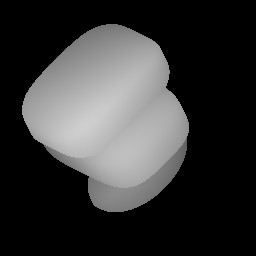}
    \end{subfigure}
    \hfill
    \begin{subfigure}[t]{\datspace}
        \includegraphics[width=\textwidth]{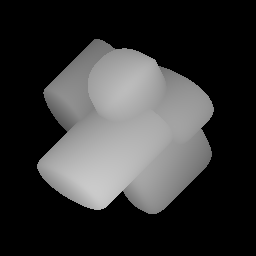}
    \end{subfigure}
    \hfill
    \begin{subfigure}[t]{\datspace}
        \includegraphics[width=\textwidth]{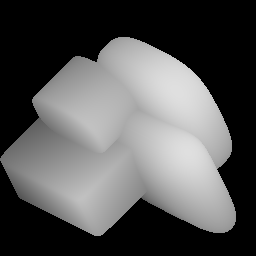}
    \end{subfigure}
    \hfill
    \begin{subfigure}[t]{\datspace}
        \includegraphics[width=\textwidth]{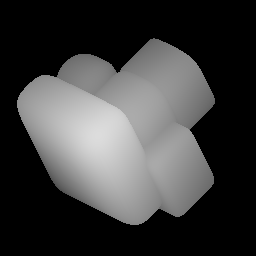}
    \end{subfigure}
    
    \begin{subfigure}[t]{\datspace}
        \includegraphics[width=\textwidth]{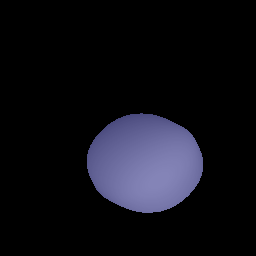}
    \end{subfigure}
    \hfill
    \begin{subfigure}[t]{\datspace}
        \includegraphics[width=\textwidth]{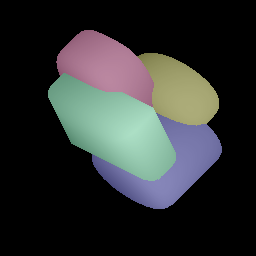}
    \end{subfigure}
    \hfill
    \begin{subfigure}[t]{\datspace}
        \includegraphics[width=\textwidth]{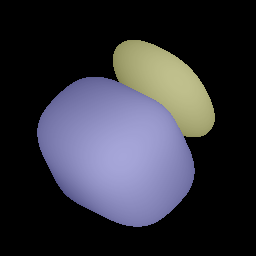}
    \end{subfigure}
    \hfill
    \begin{subfigure}[t]{\datspace}
        \includegraphics[width=\textwidth]{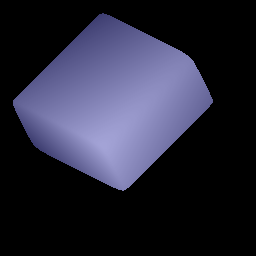}
    \end{subfigure}
    \hfill
    \begin{subfigure}[t]{\datspace}
        \includegraphics[width=\textwidth]{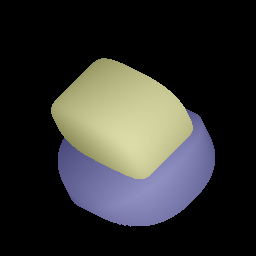}
    \end{subfigure}
    \hfill
    \begin{subfigure}[t]{\datspace}
        \includegraphics[width=\textwidth]{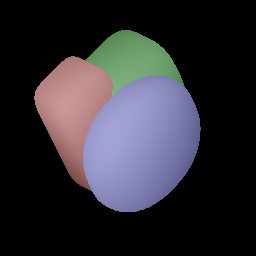}
    \end{subfigure}
    \hfill
    \begin{subfigure}[t]{\datspace}
        \includegraphics[width=\textwidth]{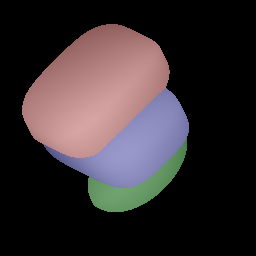}
    \end{subfigure}
    \hfill
    \begin{subfigure}[t]{\datspace}
        \includegraphics[width=\textwidth]{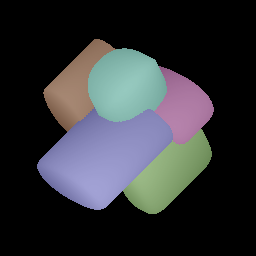}
    \end{subfigure}
    \hfill
    \begin{subfigure}[t]{\datspace}
        \includegraphics[width=\textwidth]{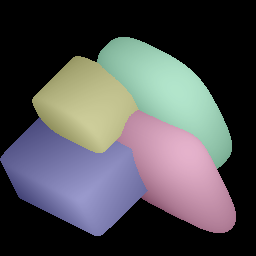}
    \end{subfigure}
    \hfill
    \begin{subfigure}[t]{\datspace}
        \includegraphics[width=\textwidth]{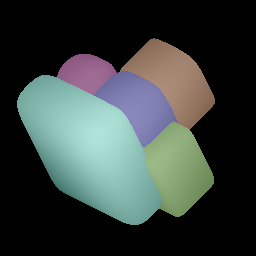}
    \end{subfigure}
    \caption{Example images from the generated dataset. The top row shows examples of the rendered images with different numbers of superquadric in the scene. The lower row shows examples of the corresponding segmentation masks. The figure is best viewed in color.}
    \label{fig:dataset_examples}
\end{figure*}

\textbf{Segment-and-fit.} Techniques from this group typically combine the segmentation and superquadric recovery stages and often rely on superquadric models to guide the segmentation~\cite{gupta1993volumetricseg,leonardis1997superquadrics,jaklic2000,horikoshi19943ddecomposition}. Due to the fact that segmentation is performed with the final scene representations (i.e., the superquadric) methods from this group are considered highly robust. However, on the down side, they often also induce a considerable computational burden on the segmentation procedure. Recently, a CNN-solution~\cite{paschalidou2019superquadrics} that falls into this group was proposed, but unlike the approach presented in this paper, was limited to segmentation of predefined classes of objects. 

\textbf{Segment-then-fit.} Techniques from this group follow a two-stage procedure, where the data is first segmented up front and independently of superquadric recovery~\cite{jaklic2000}. Thus, the entire procedure is broken down into two independent parts. Examples of techniques from this group include~\cite{gupta1990segmentation,pentland1990automatic,ferrie1993darboux,raja1994genericparts}. The solution described in this work also follow the segment-then-fit paradigm, but as we show in the experimental section result in competitive performance compared to a state-of-the-art approach from the segment-and-fit group that is in general considered to be more robust. 

\section{Dataset}\label{Sec: Dataset}

In order to train our instance segmentation and parameter estimation models, we require a large dataset of depth scenes with appropriate ground truth labels. Since no such datasets are publicly available, we generate our own and make it publicly available for the research community. In this section we present the dataset creation procedure and discuss the characteristics of the generated data.

\subsection{Prerequisites}

In this work we follow the methodology of Oblak \textit{et al.}~\cite{oblak2019recovery} and focus on unrotated superquadric models. Thus, we only try to recover the $8$ open parameters from Eq.~\eqref{eq:sq_impl} for each superquadric model and omit rotations, which introduce ambiguities in the superquadric-recovery process~\cite{oblak2019recovery}.  The main goal of this work is to extend the superquadric recovery method from~\cite{oblak2019recovery} to depth scenes with complex geometry that need to be represented with multiple superquadrics. Consequently, we fix the rotation of the objects in our dataset and render them in an axonometric projection that ensures that three sides of the objects are always visible in the rendered images. 

\subsection{Dataset creation}


We synthesize our dataset by rendering range images with multiple superquadrics in the scene. 
To construct the range images we create a custom rendering tool that accepts multiple superquadric parameter sequences. The renderer then constructs the range image of a scene by finding the surface points of the superquadrics and choosing the closest point to the viewport, if there are overlapping superquadrics in the line of sight. The scene is constrained inside a $256 \times 256 \times 256$ grid, where the first two dimensions represent the width and height of the resulting image, while the last dimension represents the depth. The scene is then mapped to the zero depth plane, resulting in a $256 \times 256$ range image, where its pixel indexes $i,j$ correspond to the $x,y$ coordinates in the 3D scene, while the pixel intensity relates to the $z$ depth in the scene.

To generate a dataset with representative superquadric objects, we uniformly sample the superquadric parameters similarly to \cite{oblak2019recovery}. However, uniformly sampling the position and size of superquadrics independently from their neighbors causes dramatic overlaps and intersections in the scene, which hides a large number of objects. We solve this by constraining the allowed intersection-over-union volume between pairs of superquadrics in each scene, where the volume is approximated using the superquadrics bounding-box. Following this requirement we first sample the number of superquadrics in the scene from the discrete uniform distribution $\mathcal{U}(1,5)$. Then, for each scene, we iteratively sample superquadric parameters. If the new superquadric intersects with the superquadrics already in the scene, we discard it and sample again. This procedure continues until there are as many superquadrics on the scene as determined in the initial sampling step. Each superquadric has its size parameters sampled from a continuous uniform distribution $\mathcal{U}(25, 76)$ and the shape parameters from $\mathcal{U}(0.01, 1)$ limiting the appearance of the rendered models to convex shapes, which are also more representative of the real world. We sample the $x_0$ and $y_0$ center coordinates from $\mathcal{U}(88, 169)$ while the $z_0$ coordinate is sampled from a tighter region $\mathcal{U}(100, 150)$. This is done to constrain the vertical overlap between the superquadrics in the scene.

Along with the range image we also render a ground truth segmentation mask image of the scene, by coloring the different visible parts of the superquadrics with a different shade of gray. This ground truth information is used for training and evaluating the segmentation model.

\subsection{Dataset totals}

The complete dataset contains 120000 rendered scenes and corresponding segmentation masks. We also store range images of individual superquadrics in each scene in the dataset along with their parameters. For the experiments we split the dataset into three disjoint parts: for training, validation and testing. We use the training set to learn the parameters of our models, the validation set to observe over-fitting issues during training and the test for the final performance evaluation. A few illustrative examples from the generated dataset together with the corresponding segmentation masks are shown in Fig.~\ref{fig:dataset_examples} and a high-level summary of the dataset and experimental setting is given in Table~\ref{tab:dataset_characteristics}.

\begin{table}[t]
    \centering
    \caption{Dataset summary.}
    \resizebox{\columnwidth}{!}{%
    \begin{tabular}{l r r r r r r}
    \hline
    \hline
    \#Superquadrics & $1$ & $2$ & $3$ & $4$ & $5$ & Any\\
    \hline
    \#Train Images & $15882$ & $16108$ & $15930$ & $15983$ & $16097$ & 80000\\
    \#Validation Images & $3989$ & $3944$ & $4020$ & $3948$ & $4099$ & $20000$ \\
    \#Test Images & $3949$ & $4023$ & $3996$ & $4059$ & $3973$ & $20000$ \\
    \hline
    \hline
    \end{tabular}}
    \label{tab:dataset_characteristics}
\end{table}

\section{Superquadric recovery methodology}\label{Sec: methods}

In this section we now present our approach to  segmentation and recovery of multiple superquadrics using CNN models. 

\subsection{Segmentation}

As our range images contain multiple objects of the same class (i.e., superquadric-like objects), we resort to instance segmentation to identify parts of the range images belonging to structures that can be represented with superquadrics. One of the most popular models for instance segmentation is Mask R-CNN \cite{MaskRCNN}, which operates in a two-stage fashion. In the first stage, it uses a region proposal network (RPN) that finds candidate regions in the image. In the second stage, the final predictions are made. Here, three model heads are used: one for detection (two-class classification: object present or not), one for regression of the bounding boxes, and one for prediction of the binary segmentation mask.

In our implementation, we use a ResNet-$101$ \cite{resnet} backbone as the feature extractor along with a feature pyramid network (FPN) that makes it possible to exploit multiple scales of the feature maps. These features get fed trough a region proposal network which predicts object scores and their bounding boxes at each feature position. The predictions are then filtered by a non-maximum suppression algorithm, which removes overlapping bounding boxes.

The RPN bounding boxes and the FPN features get combined using the RoIAlign operator and fed into the three network heads to obtain the final class (object present or not), bounding box, and binary mask for each region proposal. Here the classification scores are used for the elimination of any background instances.
For more information on Mask R-CNNs, the reader is referred to  \cite{R-CNN_Girshick, FastRCNN_Girshick_2015_ICCV, DBLP:journals/corr/RenHG015, FPN, MaskRCNN}.


\begin{table}[t]
    \centering
    \caption{Architecture of the CNN regressor used for superquadric parameter estimation. 
    }
    \resizebox{\columnwidth}{!}{%
    \begin{tabular}{l r r r }
    \hline
    \hline
    \# & Output size & Layer operation & \#kernels, size, stride \\
    \hline
    1 & $128\times 128$ & Conv2D+BN+ReLU & $32, \, 7 \times 7 , \, s 2$ \\
    2 & $128\times 128$ & Conv2D+BN+ReLU & $32, \, 3 \times 3 , \, s 1$ \\
    3 & $128\times 128$ & Conv2D+BN+ReLU & $32, \, 3 \times 3 , \, s 1$ \\
    
    4 & $64\times 64$ & Conv2D+BN+ReLU & $32, \, 3 \times 3 , \, s 2$ \\
    5 & $64\times 64$ & Conv2D+BN+ReLU & $64, \, 3 \times 3 , \, s 1$ \\
    6 & $64\times 64$ & Conv2D+BN+ReLU & $64, \, 3 \times 3 , \, s 1$ \\
    
    7 & $32\times 32$ & Conv2D+BN+ReLU & $64, \, 3 \times 3 , \, s 2$ \\
    8 & $32\times 32$ & Conv2D+BN+ReLU & $128, \, 3 \times 3 , \, s 1$ \\
    9 & $32\times 32$ & Conv2D+BN+ReLU & $128, \, 3 \times 3 , \, s 1$ \\
    
    10 & $16\times 16$ & Conv2D+BN+ReLU & $128, \, 3 \times 3 , \, s 2$ \\
    11 & $16\times 16$ & Conv2D+BN+ReLU & $256, \, 3 \times 3 , \, s 1$ \\
    12 & $16\times 16$ & Conv2D+BN+ReLU & $256, \, 3 \times 3 , \, s 1$ \\
    13 & $8\times 8$ & Conv2D+BN+ReLU & $256, \, 3 \times 3 , \, s 2$ \\
    14 & $16384$ & Flatten & N/A\\
    15 & $8$ & Dense & N/A \\
    \hline
    \hline
    \end{tabular}}
    
    \label{tab:architecture}
\end{table}

\begin{figure*}[tb]
    \centering
    \begin{subfigure}[t]{0.104\linewidth}
        \includegraphics[width=\textwidth]{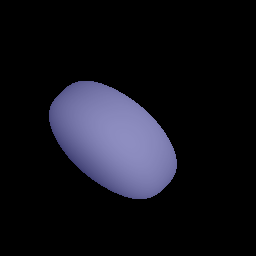}
    \end{subfigure}
    \begin{subfigure}[t]{0.104\linewidth}
        \includegraphics[width=\textwidth]{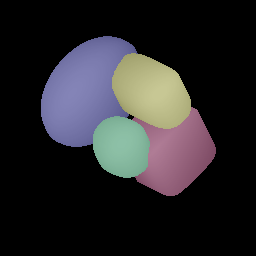}
    \end{subfigure}
    \begin{subfigure}[t]{0.104\linewidth}
        \includegraphics[width=\textwidth]{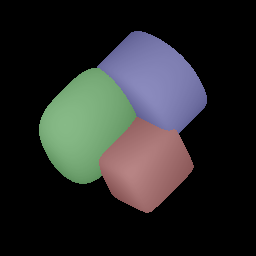}
    \end{subfigure}
    \hfill
    \begin{subfigure}[t]{0.104\linewidth}
        \includegraphics[width=\textwidth]{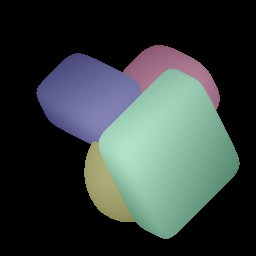}
    \end{subfigure}
    \begin{subfigure}[t]{0.104\linewidth}
        \includegraphics[width=\textwidth]{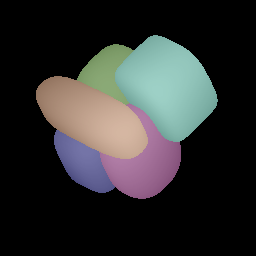}
    \end{subfigure}
    \begin{subfigure}[t]{0.104\linewidth}
        \includegraphics[width=\textwidth]{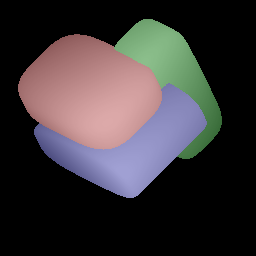}
    \end{subfigure}
    \hfill
    \begin{subfigure}[t]{0.104\linewidth}
        \includegraphics[width=\textwidth]{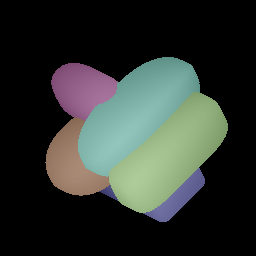}
    \end{subfigure}
    \begin{subfigure}[t]{0.104\linewidth}
        \includegraphics[width=\textwidth]{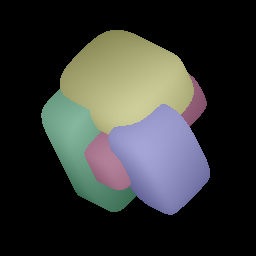}
    \end{subfigure}
    \begin{subfigure}[t]{0.104\linewidth}
        \includegraphics[width=\textwidth]{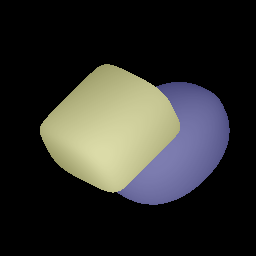}
    \end{subfigure}
    
    \begin{subfigure}[t]{0.104\linewidth}
        \includegraphics[width=\textwidth]{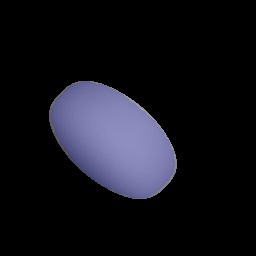}
        \caption{100\%}
    \end{subfigure}
    \begin{subfigure}[t]{0.104\linewidth}
        \includegraphics[width=\textwidth]{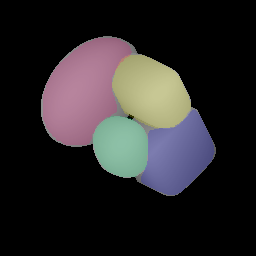}
        \caption{93.8\%}
    \end{subfigure}
    \begin{subfigure}[t]{0.104\linewidth}
        \includegraphics[width=\textwidth]{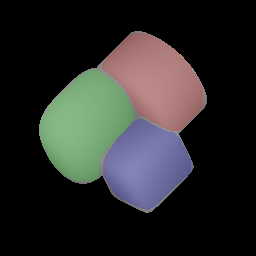}
        \caption{93.3\%}
    \end{subfigure}
    \hfill
    \begin{subfigure}[t]{0.104\linewidth}
        \includegraphics[width=\textwidth]{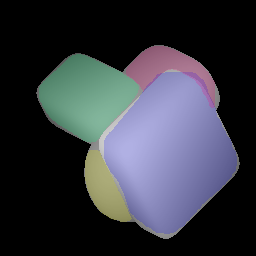}
        \caption{86.9\%}
    \end{subfigure}
    \begin{subfigure}[t]{0.104\linewidth}
        \includegraphics[width=\textwidth]{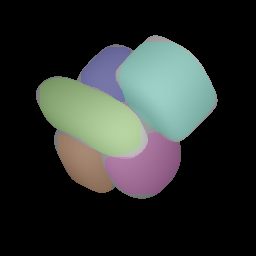}
        \caption{80\%}
        \label{subfig:seg93}
    \end{subfigure}
    \begin{subfigure}[t]{0.104\linewidth}
        \includegraphics[width=\textwidth]{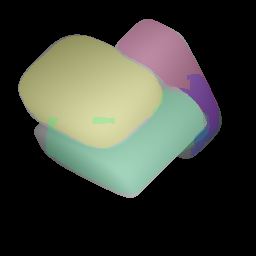}
        \caption{76.7\%}
        \label{subfig:seg7}
    \end{subfigure}
    \hfill
    \begin{subfigure}[t]{0.104\linewidth}
        \includegraphics[width=\textwidth]{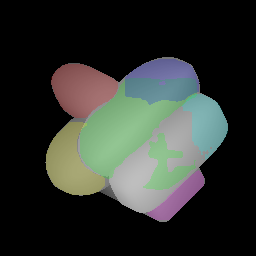}
        \caption{55.7\%}
    \end{subfigure}
    \begin{subfigure}[t]{0.104\linewidth}
        \includegraphics[width=\textwidth]{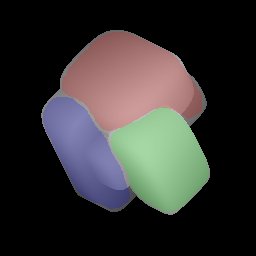}
        \caption{52.5\%}
        \label{subfig:seg14}
    \end{subfigure}
    \begin{subfigure}[t]{0.104\linewidth}
        \includegraphics[width=\textwidth]{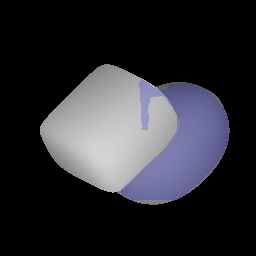}
        \caption{40\%}
    \end{subfigure}
    \caption{Predicted segmentation masks from the Mask R-CNN model. The images are ordered in columns of three. Three good predictions (left), three average predictions (middle) and three bad predictions (right). In the first row we show range images with overlaid ground truth masks. The second row shows masks obtained with our segmentation model. Under the images we also report the mAP value for the segmentation. Most of the predictions are sufficient, even in the average subsection of the predictions. We observe that fine details are elusive to the model, such as disconnected masks (h) or narrow subparts of masks (e,f). Best viewed in color.}
    \label{fig:segmentations}
\end{figure*}

\subsection{Parameter estimation}

Once the scene is segmented and superquadric-like objects are identified in the input images, we feed the predictions into a CNN regressor for parameter estimation. We follow the work of \cite{oblak2019recovery} and use a regression model derived from the popular VGG architecture \cite{parkhi2015deep}. The model is designed as a $13$ layer CNN with a fully-connected layer of size $8$ on top. Each conv layer is followed by batch normalization and a ReLU activation, which reduces overfitting and allows the model to better generalize. The model is summarized in Table~\ref{tab:architecture}. 

The input to the CNN regressor is a range image containing a single superquadric-like instance and the output is a prediction of $8$ parameters describing the size, shape and position, of the superquadric representing the input data, i.e., $\mathbf{y} = [a_1, a_2, a_3, \varepsilon_1, \varepsilon_2, x_0, y_0, z_0]$. Different from \cite{oblak2019recovery}, the inputs to our model are not necessarily complete superquadrics, but automatically segmented range data, where parts of the object may be occluded due to overlap with other objects in the scene. Thus, we account for this in our training procedure and learn the parameters of our regressor by utilizing occluded data. As we show in the experimental section this allows us to quite efficiently estimate superquadric parameters even if part of the data is missing either due to occlusions or errors in the segmentation steps.      


\section{Experiments and results}\label{Sec: experiments and results}


\subsection{Instance segmentation}

The Mask R-CNN backbone is initialized with a ResNet-$101$ structure \cite{resnet}, pre-trained on the MS COCO dataset \cite{MSCOCO}. The training is split into two stages. In the first stage, we lock the training of the backbone and set the learning rate to $10^{-3}$, with momentum of $0.9$. In the second stage we unlock the backbone and fine-tune the network with a smaller learning rate of $10^{-4}$. We present the standard mean average precision (mAP) scores of the instance segmentation in \Cref{tab:AP}, as used in the COCO challenge. The model is trained on $80k$ training range-images of superquadric scenes, with a batch size of $2$. We use an additional $20k$ images for validation and $20k$ images for testing. 
The model is trained on an NVIDIA GTX TITAN X GPU. 

In Table~\ref{tab:AP} we report the segmentation results using our Mask R-CNN model. We can see that average precision at Intersection-over-Union (IoU) thresholds $50\%$ and $75\%$ are higher than the averaged mAP over multiple IoU thresholds. This indicates that the model fail only at the highest intersections, segmenting the objects with good detail and precision.

\begin{table}[tb]
    \caption{Instance segmentation results. mAP$_{50}$ and mAP$_{75}$ denote scores computed at $50\%$ and $75\%$ IoU respectively, while mAP denotes the mean average precision averaged over IoU values from $50\%$ up to $95\%$, taken at $5\%$ steps.}
    \centering
    \small
    \begin{tabular}{r r r}
        \hline
        \hline
        mAP & mAP$_{50}$ & mAP$_{75}$ \\
        \hline
        85.57 & 97.33 & 95.95 \\
        \hline
        \hline
    \end{tabular}
    \label{tab:AP}
    \vspace{-3mm}
\end{table}

In Figure~\ref{fig:segmentations} we present some examples of predicted masks for the training set. Most of the objects have been segmented with sufficient precision. On average, the model only misses smaller and highly occluded objects (\Cref{subfig:seg93,subfig:seg7}). It also struggles with objects visually cut in half because of overlaps (\Cref{subfig:seg14}). 
In these cases we either get multiple separate instance segments or the model fails to detect one of the parts completely. We suspect this might be caused by significant bounding box overlap between the foreground and background objects. The latter causing the former to get suppressed by the Mask R-CNN non maximum suppression algorithm.

\subsection{Parameter prediction}

\begin{table*}[t]
\caption{Parameter-prediction performance. The table shows MAE scores for each of the $8$ superquadric parameters. The rows show results on different subsets of segmented range images test set, defined by the number of superquadrics the parent scene. The ``All'' row shows scores averaged over the entire set.}
\vspace{-0.35cm}
\begin{center}
\small
\begin{tabular}{l c c c c c c c c c c c}
\hline\hline
\multirow{ 2}{*}{\#sq} & & \multicolumn{3}{c}{Dimensions [0-256]} & & \multicolumn{3}{c}{Position [0-256]} & & \multicolumn{2}{c}{Shape [0-1]} \\
\cline{3-5} \cline{7-9} \cline{11-12} 
 & & $a_1$ & $a_2$ & $a_3$ & &  $x_0$ & $y_0$ & $z_0$ & &  $\epsilon_1$ & $\epsilon_2$ \\
\hline
All & & $1.134$ & $1.187$ & $1.248$  & & $1.953$ & $1.864$ & $2.639$  && $0.017$ & $0.017$ \\
\hline
1 & & $0.515$ & $0.555$ & $0.537$  & & $0.957$ & $0.925$ & $2.154$  && $0.009$ & $0.008$ \\
2 & & $0.681$ & $0.736$ & $0.728$  & & $1.165$ & $1.093$ & $2.181$  && $0.011$ & $0.010$ \\
3 & & $0.930$ & $0.984$ & $1.036$  & & $1.528$ & $1.448$ & $2.386$  && $0.013$ & $0.013$ \\
4 & & $1.580$ & $1.646$ & $1.708$  & & $3.066$ & $2.966$ & $3.110$  && $0.026$ & $0.025$ \\
5 & & $1.201$ & $1.241$ & $1.357$  & & $1.776$ & $1.669$ & $2.685$  && $0.017$ & $0.017$ \\
\hline\hline
\end{tabular}
\label{tab:MAE}
\end{center}
\vspace{-6mm}
\end{table*}

We initialize the parameter prediction model with the weights from \cite{oblak2019recovery}, as the same neural network architecture was used in that work. To train the parameters of the model we use the ADAM minibatch stochastic gradient descent optimisation algorithm, which minimizes the MSE loss. We set the learning rate of the algorithm to $10^{-3}$ and keep the rate constant during training. As already indicated above, we use the segmentations produced by our Mask R-CNN model as the basis for the training to make the model robust to missing data. 
We only train on segmentations with an IoU higher than $50\%$ compared to the ground truth masks. The model is trained for $63$ epochs, with varying batch sizes constructed always from batches of $4$ scene range images, giving us a maximum batch size of $20$ segmented range images. We report performance for the CNN regressor in terms of the Mean Absolute Error (MAE) between the predicted and ground truth  parameters. This measure was sufficient for our problem, since we predict superquadric parameters for superquadric visualizations, where the matching of parameters correlates with the 3D matching of the objects.


In \Cref{tab:MAE} we present the MAE scores for each parameter on a test set of $20000$ images. 
In addition to the MAE score for the entire test set, we also show separate MAE scores for scenes with different numbers of superquadrics. On average the model performs very well, predicting position and size in the order of one pixel accuracy compared to the $[0,256]$ range of possible values. The shape parameters $\varepsilon_0$ and $\varepsilon_1$ also achieve about 0.017 mean absolute error which is also small compared to the $[0, 1]$ range of possible values. The model performs better in scenes with a smaller number of superquadrics since more superquadrics in the scene typically result in greater intersections and occlusions. \Cref{tab:MAE} shows an almost monotonous increase in MAE as the number of superquadrics is increased, the only disparity is a larger error in scenes with $4$ objects than in scenes with $5$.

\begin{figure}[t!]
    \centering
    \includegraphics[width=\linewidth]{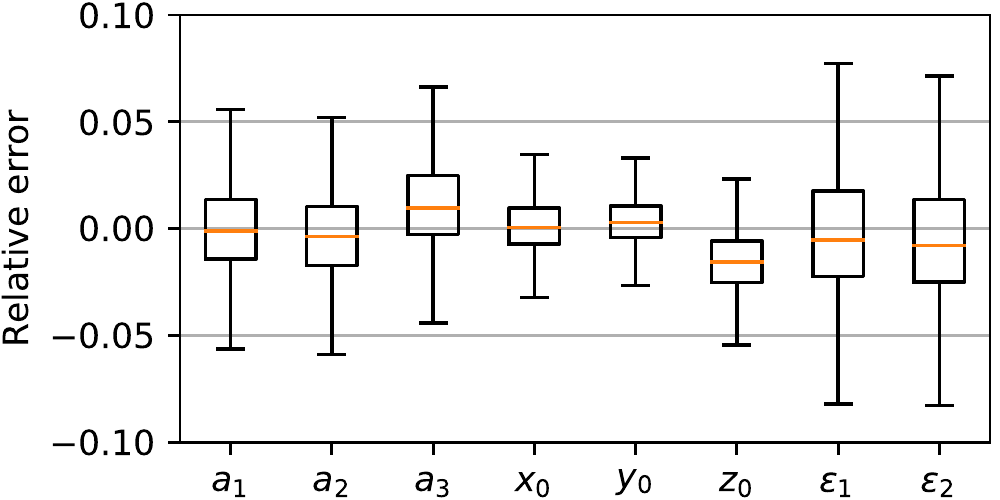}
    \caption{Box-and-whiskers plots of the relative error for each parameter.}
    \label{fig:boxplot1}
    \vspace{-0.3cm}
\end{figure}

In \Cref{fig:boxplot1} we show box-and-whiskers plots of the relative errors between ground truth and the predicted parameter values over the entire test set of segmented range images. We see that most of the error mass is close to the mean. The positional parameters are predicted with especially small variance in their errors. We also observe that the $z$ axis size parameters are on average slightly overestimated. This seems to get compensated by an underestimation of the $z$ axis position, thus aligning the top surface of the ground truth and the predicted superquadrics.

\begin{figure*}[t!]
    \centering
    \includegraphics[width=0.9\linewidth]{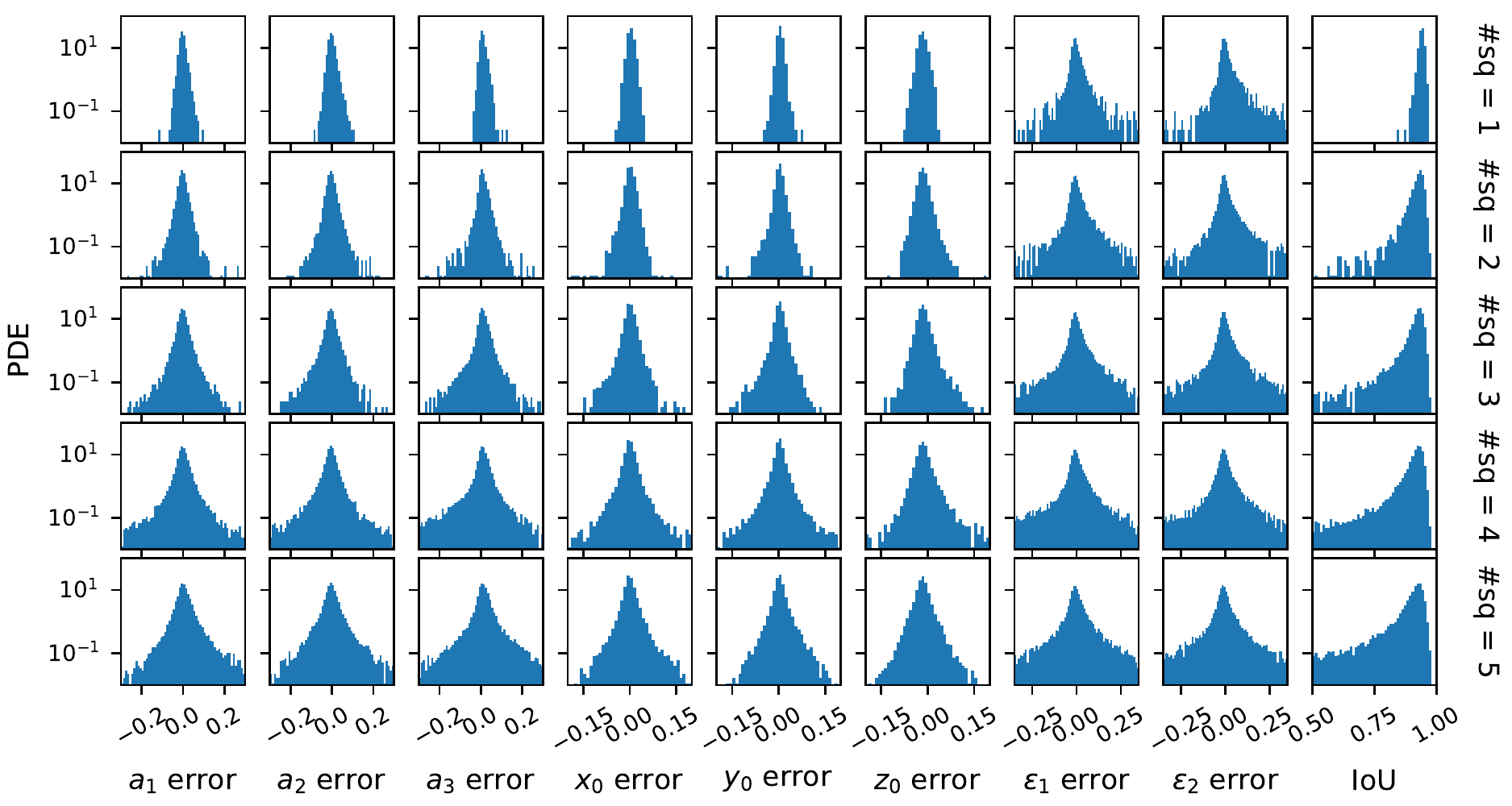}
    \vspace{-2mm}
    \caption{Our methods error distribution for each parameter. Each row shows results obtained from the 5 subsets scene images, each with a different number of superquadrics in its scenes. We also add the last column showing the IoU distribution of the predicted masks with Mask R-CNN.}
    \label{fig:heavy_tails}
\end{figure*}
\begin{figure*}[htb]
    \newcommand{\syspace}{0.10\linewidth}
    \centering
    \begin{subfigure}[t]{\syspace}
        \includegraphics[width=\textwidth]{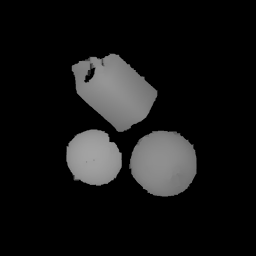}
    \end{subfigure}
    \quad
    \begin{subfigure}[t]{\syspace}
        \includegraphics[width=\textwidth]{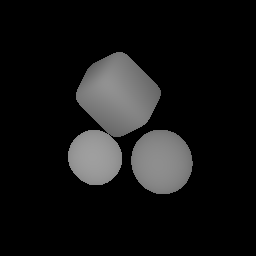}
    \end{subfigure}
    \begin{subfigure}[t]{\syspace}
        \includegraphics[width=\textwidth]{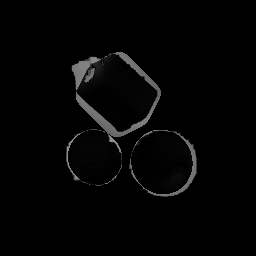}
    \end{subfigure}
    \quad
    \begin{subfigure}[t]{\syspace}
        \includegraphics[width=\textwidth]{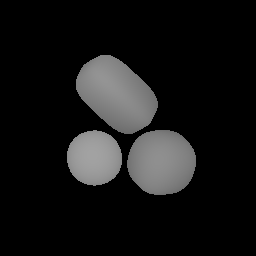}
    \end{subfigure}
    \begin{subfigure}[t]{\syspace}
        \includegraphics[width=\textwidth]{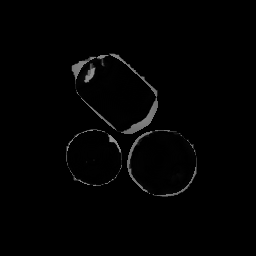}
    \end{subfigure}
    
    \begin{subfigure}[t]{\syspace}
        \includegraphics[width=\textwidth]{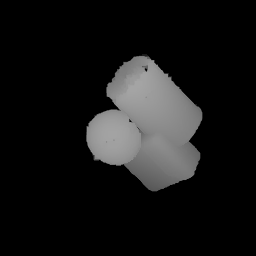}
    \end{subfigure}
    \quad
    \begin{subfigure}[t]{\syspace}
        \includegraphics[width=\textwidth]{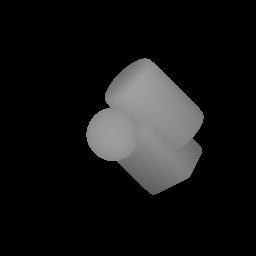}
    \end{subfigure}
    \begin{subfigure}[t]{\syspace}
        \includegraphics[width=\textwidth]{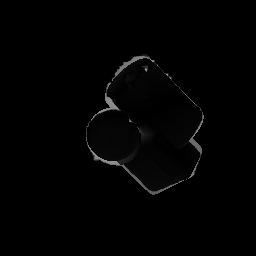}
    \end{subfigure}
    \quad
    \begin{subfigure}[t]{\syspace}
        \includegraphics[width=\textwidth]{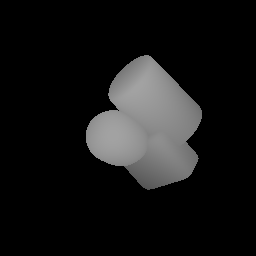}
    \end{subfigure}
    \begin{subfigure}[t]{\syspace}
        \includegraphics[width=\textwidth]{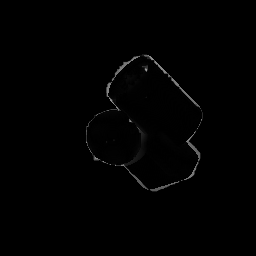}
    \end{subfigure}
    
    \begin{subfigure}[t]{\syspace}
        \includegraphics[width=\textwidth]{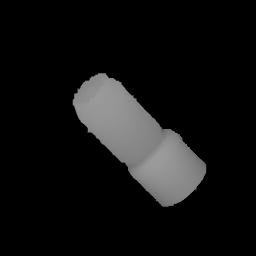}
    \end{subfigure}
    \quad
    \begin{subfigure}[t]{\syspace}
        \includegraphics[width=\textwidth]{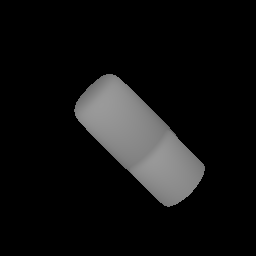}
    \end{subfigure}
    \begin{subfigure}[t]{\syspace}
        \includegraphics[width=\textwidth]{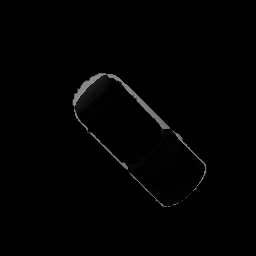}
    \end{subfigure}
    \quad
    \begin{subfigure}[t]{\syspace}
        \includegraphics[width=\textwidth]{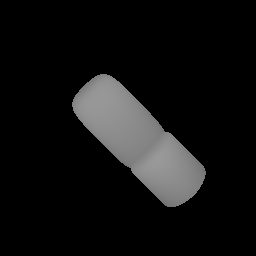}
    \end{subfigure}
    \begin{subfigure}[t]{\syspace}
        \includegraphics[width=\textwidth]{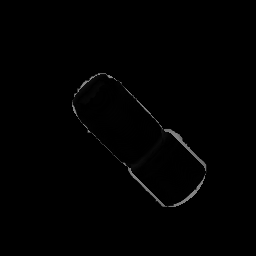}
    \end{subfigure}
    \caption{Qualitative comparison with the state-of-the-art: Input range images of (scanned) real-world objects (first column), Our reconstructions (second column), Absolute difference between the ground truth and our reconstruction (third column), Reconstructions by Leonardis et.\ al.~\cite{leonardis1997superquadrics, jaklic2000} (fourth column), Absolute difference between the ground truth and reconstruction by Leonardis et.\ al.~\cite{leonardis1997superquadrics, jaklic2000} (last column). 
    }
    \label{fig:real_ims}
    \vspace{-3mm}
\end{figure*}

Scenes with larger numbers of superquadrics are harder to segment, occasionally giving our parameter prediction model highly corrupted segmentation masks, that can either blend range information from multiple objects into one segmented range image or return smaller subsets of the actual masks. On such corrupt segmented range images our prediction model naturally performs much worse than on cleaner segmentations, resulting in a somewhat heavy-tailed error distribution. We show this in \Cref{fig:heavy_tails} where we plot the error distribution for all parameter predictions and subsets over the number of superquadrics in the scene. We also show how our segmentation model performs on each subset by showing the distribution of IoU values for its predicted segmentations. The distributions move away from a Gaussian shape quickly when more than one superquadric is present in the scene. The tails become larger when we increase the number of objects in the scene. As mentioned earlier, this can be explained by the inefficiency of the segmentation model, as the model also performs worse with greater numbers of objects in the scene - the  IoU distribution becomes more and more skewed, with a heavier tail.

We also compare our approach to the state-of-the-art segmentation and superquadric recovery method from~\cite{leonardis1997superquadrics, jaklic2000} on range-images of real objects.
For this experiment, we used range-image scans of real objects taken by Oblak et.\ al.\ for their work in \cite{oblak2019recovery}. We constructed range image scenes of multiple object by shifting the original images in pixel space and combining them using the $\max$ operator.
The original range images, and their superquadric reconstructions using our approach and the state-of-the-art method from~\cite{leonardis1997superquadrics, jaklic2000} are shown in \Cref{fig:real_ims}. The iterative method from~\cite{leonardis1997superquadrics, jaklic2000} performs comparably to our solution, as we can see from the examples. Our method achieved $2.79$ MAE calculated over all pixels differences from all pairs of ground truth and reconstructed images while~\cite{leonardis1997superquadrics, jaklic2000} scored $1.78$. However, we  note that the iterative algorithm of the original method  results in much higher processing times. Our method performs similarly in terms of reconstruction quality, but  computes the segmentations and parameter predictions with a $100\times$ speed up over the state-of-the-art approach. Specifically, the iterative method converges in about $10$ s on one image while our method needs $0.11$ s on a GPU. While our methods advantage against~\cite{leonardis1997superquadrics, jaklic2000} is that we can parallelize its computations, it still performs faster on a single threaded CPU with about $5$ s per image.


\section{Conclusion}\label{Sec: Conclusions}

We have presented a CNN-based solution for segmentation and recovery of multiple superquadrics from range images. We have shown that the designed solution is able to efficiently decompose complex depth scenes into smaller parts that can be modelled by superquadric models. Our approach was shown to produce scene reconstruction on par with a state-of-the-art method from the literature, while ensuring a significant speed up in processing times. As part of our future work, we will extend the solution to account for rotated superquadrics as well.  

\section*{Acknowledgements}

{\footnotesize This research was supported in parts by the ARRS (Slovenian Research Agency) Project J2-9228 
``A neural network solution to segmentation and recovery of superquadric models from 3D image data'',
ARRS Research Program P2-0250 (B) ``Metrology and Biometric Systems'' and the 
ARRS Research Program P2-0214 (A) ``Computer Vision''. }

{
\footnotesize	
\bibliographystyle{ieee}
\bibliography{bibs_short}
}
\clearpage

\end{document}